\documentclass[runningheads,a4paper]{llncs}
\usepackage{amssymb}
\usepackage{amsmath}
\usepackage{graphicx}

\usepackage{url}
\urldef{\mailsa}\path|{gordeev-d-i, rkpotapova}@yandex.com|
\newcommand{\keywords}[1]{\par\addvspace\baselineskip
\noindent\keywordname\enspace\ignorespaces#1}

\usepackage{graphicx}
\begin{document}
\bibliographystyle{splncs03}

\mainmatter  

\title{Detecting state of aggression in sentences using CNN}	

%
%
\author{Rodmonga Potapova%
\and Denis Gordeev}
\authorrunning{R. Potapova, Detecting state of aggression in sentences using CNN}

\institute{Moscow State Linguistic University,\\
Ostozhenka. 38, 119034 Moscow, Russia\\
\mailsa\\
}

%
%

\maketitle

\begin{abstract}
In this article we study verbal expression of aggression and its detection using machine learning and neural networks methods. We test our results using our corpora of messages from anonymous imageboards. We also compare Random forest classifier with convolutional neural network for "Movie reviews with one sentence per review" corpus.
\end{abstract}

\section{Introduction}
With the development of the Internet, verbal aggression and cyberbullying have become a problem on the Net. For example, the US government has proposed an incentive to stop cyberbullying \cite{url:cyberbullying} and Russian criminal code \cite{book:ruscrimcode} persecutes verbal aggression both in oral and written speech, but there is no clear definition of what is aggression and what is not. For this reason many researchers study expression of aggression both on all levels nowadays. A lot of studies have been devoted to the analysis of the semantic field of aggression \cite{lncschap:potapova1}⁠ and its aspects in multicultural communities both for written  \cite{lncschap:potapova2,lncschap:potapovakomalova} and oral speech \cite{lncschap:potapovabobrov,lncschap:potapovapotapov}.

In recent years the number of works using neural network and machine learning techniques has seen a dramatic increase. While neural networks per se or combined with machine learning methods such as Random forest \cite{book:randomforest}⁠ or SVM \cite{proceeding:svm}⁠  make it possible to discard vocabulary-based methods and to switch from these and other similar methods requiring manual editing. That is why the field of sentiment analysis, being already studied \cite{lncschap:kharlamov2014}, has become even more prominent with the advent of machine learning and neural networks as it can be efficiently solved using these methods. Skip-gram \cite{jour:skipgram}, glove \cite{proceeding:glove}⁠ and other word vector models have shown better results than usual bigram and trigram models and helped to overcome the computational difficulties of larger n-gram spaces. They provided an efficient and computationally affordable method of finding similarity between different words and building semantic vector space. They require no annual annotation, only large corpora of texts, thus any set of texts can be used as a corpus. Skip-gram models combined with deep learning methods are widely used now for sentiment analysis \cite{proceeding:santos}⁠, for object labeling \cite{jour:lazaridou}  and for other NLP tasks.

CNN neural networks first used for image object classification and detection and other computer vision tasks, have been shown to be efficient for NLP tasks. Kim \cite{jour:kim} has proposed to combine CNN's and Word2vec for sentiment analysis tasks. His model outperformed other deep learning models for the majority of tasks on different corpora. His model included word2vec \cite{jour:skipgram} word embeddings for every word of the text and a set of convolutional filters. Chunting Zhou et. all \cite{jour:zhou} have showed close results to Kim's CNN using a LSTM-model. It was also \cite{url:rakhlin}⁠ proposed to decrease the size and number of filters for such a tiny corpus as "Movie reviews with one sentence per review" \cite{proceeding:panglee}.

In this article we compare the results of a CNN-model with usual machine learning techniques (Random forest) for the task of analysis of aggression. 

We used a corpus of movie reviews by Pang and Lee and our tiny corpus of aggressive imageboard messages for this task and compared the results with our Random Forest classifier.

\section{Methods and Materials}
We selected imageboards (4chan.org, 2ch.hk) as the material for our tiny corpus because these communities are considered to be extremely aggressive and messages containing expression of verbal aggression are abundant there \cite{lncschap:potapovagordeev}. Bernstein who has conducted research on imageboard culture supports this statement \cite{proceeding:bernstein}. By verbal aggression we understand a “personality trait that predisposes persons to attack the self‐concepts of other people instead of, or in addition to, their positions on topics of communication \cite{book:infante}.⁠” 

Movie reviews corpus is a subset of Stanford Sentiment Treebank containing only one-sentence reviews. Data is labeled there as postive or negative (Table 1).

Imageboard aggression corpus of English messages consists of about 2000 annotated  messages for Russian and English languages. Both parts consist of about 1000 messages. They are labeled as positive or negative, neutral reviews are removed. There is no test data, so 90\% of data was used training and 10\% for evaluation (see Table \ref{table:corpora}). From the table we can see that the vocabulary is much more diverse for movie reviews, while imageboards suffer from primitive lexics.

\begin{table}
\caption{Comparison of Movie reviews and Aggressive messages corpora}
\begin{center}
\begin{tabular}{l*{6}{c}r}
Corpora   & Classes (N) & Avg. sentence length & Dataset (sent.) & Voc. size & Voc. in model\\
\hline
Movie reviews & 2 & 20 & 10662 & 18765 & 17121 \\ 
SVAggr. (eng.) & 2 & 19   & 19732  & 3765  & 3690 \\
SVAggr. (rus.) & 2 & 13 & 5101 & 1030 & 989 \\

\end{tabular}
\end{center}
\textit{Movie reviews is a corpus of on sentence per movie review, SVAggr.(eng.) is the corpus of American anonymous message, SVAggr.(rus) is the corpus of Russian anonymous message}
\label{table:corpora}

\end{table}

We used a CNN-non-static model similar to Kim's with some adjustments suggested by Rakhlin. We decreased the number of filter sized from 3 to 2 and  However, instead of using Google-news corpus or training word2vec vectors we used model based on 4chan.org threads with about 600-dimensional vector trained on about 1,089,000 messages and containing about 30 million words for the American corpus. For 2ch.hk and the Russian language we used our corpus of 974654 messages containing 13640000 sentences, For the task of aggression classification we treated messages as single sentences. There is also no test data, and 90\% of data was used training and 10\% for evaluation.

For the Random forest classifier, first, we chose a set of words and phrases using our background knowledge of typical expressions for aggressive messages. Then after clearing and tokenizing raw data we computed features $ F $ used for Random forest training and evaluation.
\begin{equation}
   F_{1,i} = \{f_{1}\, f_{2}, ..., f_{i}\}
\end{equation}
where $ \{f_{1}, f_{2}, ..., f_{i}\} $ is a set features for sentences $ \{ s_{1}, s_{2}, ..., s_{i} \} $ and 
\begin{equation}
\begin{array}{rcl}
\dot f{n} = \{\sum_{i=1}^{z} (w2v (\{ w_{1},w_{2}, ..., w_{z} \}) ), mean(w2v(s_{n})),  \\ max(w2v(s_{n})) - min(w2v(s_{n})), len(s_{n})) \}
\end{array}
\end{equation}

where $ s{n} $ is a sentence with words $ \{w_{1}, w_{2}, ..., w_{z}\} $ and $ w2v $ is a function that computes distance between a given word and a chosen set of aggressive words and phrases for sentence $ s_{n} $, len is a length of the sentence $ s{n} $. 

\begin{figure}[htp]
\centering
\includegraphics[scale=0.4]{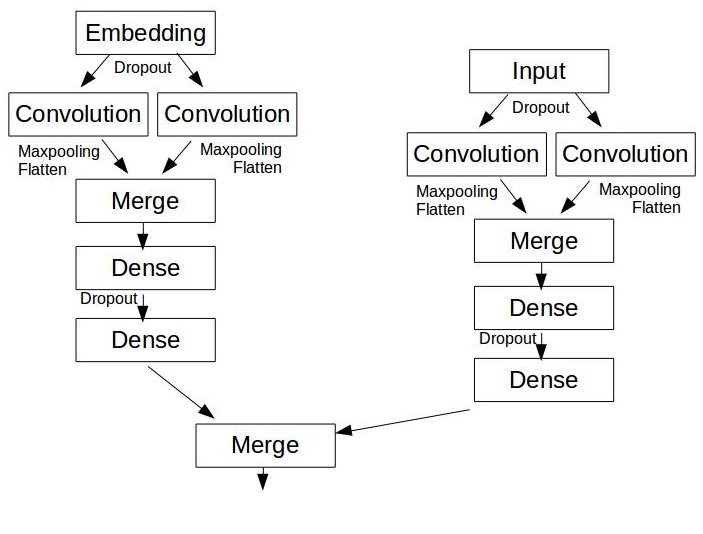}
\caption{The structure of neural networks. \textit{On the right is a branch using POS-tags, on the left is CNN with word-embedding vectors.}}
\label{figure}
\end{figure}

We also tried to use some language features to improve CNN-predictions. As linguistic information we used preprocessed part-of-speech (POS) tags. We tried to implement these tags into a neural model, so we added another neural model in parallel and then merged its results with the CNN-neural network that used word embeddings into a final neural network. We tried several models for the POS neural network. At first we tried usual recurrent network, it gained decent 76 \% after the 5th epoch for the aggression detection corpus,however, it overfitted soon after it. Then we tried the same model as used for the word2vec embeddings. The only difference was that we changed word2vec embeddings with random coefficients, it helped to get decent 81.1 \% for the same task, almost the same as with using word2vec embeddings. After that, we tried to combine two models, unfortunately, we increased the results above the threshold of CNN-rand model that used nothing besides part-of-speech tags only a little (see Figure \ref{figure}).

\section{Results and discussions.}
Results of CNN-based and Random forest models are listed in Table 2.

Random Forest classifier outperforms CNN-based methods for aggression detection task for the English language. It can be a result of overfitting, although train and test sets are not mixed and properly divided.  Also it is possible to consider that word2vec distance performs well enough for the task of aggression detection and that features were selected successfully or that concatenating sentences is not effective. Moreover, with the increase of the set, the results are prone to worsen because more types of aggression will be included and it will be expressed in other wording. We can see that it performed purely for movie reviews classification task, also we failed to select a good set of feature words. Moreover, as said by Chunting Zhou [8]⁠ a simple SVM algorithm with hand-crafted features outperformed more robust and complicated models, however, it requires manual featuring.

We also supported the results of the article by \cite{url:rakhlin}. It is asserted there that the result of the work by Kim is caused not by the amount and complexity of convolutional layers. So Kim's model may be greatly simplified without affecting the performance. We should also admit that Kim says himself that the philosophy of his work is that pretrained deep learning features work well for other tasks as well and asserts it in conclusion as well.

\begin{table}
\label{tab:tab2}
\caption{Results of CNN and Random forest classificators}
\begin{center}
\begin{tabular}{l*{6}{c}r}
Classifier   & MR (\%) &   Verb. Aggr. (eng.) (\%) & Verb. Aggr. (rus.)\\
\hline
Random Forest & 58.39 & 88.4  & 59.13\\
CNN-non-static	& 81.1 & 81.39  & 66.68\\
CNN-rand (POS)  & 80.9 & 81.17   & 62.37\\ 
CNN-non-static, CNN-rand (POS) combined & 81 & 81.22 & 64.53
\end{tabular}
\end{center}
\textit{MR - Movie reviews corpus, Verb. aggr. (eng.)- is a corpus of American imageboard messages annotated with consideration of containing or not state of aggression, Verb. Aggr. (rus.) is the corpus of Russian anonymous messages}
\end{table}

Also using not a Google news word2vec model, but a model from another domain having substantial vocabulary did not affect the results

\section{Conclusion}
In this article we have considered ways of automatic determining of state of aggression. We used two classifiers for this task and later compared them. We used Random forest classifier and a convolutional neural network for this task. They were tested on two different corpora: "Movie reviews with one sentence per review" containing positive and negative movie reviews and a 2-language Anonymous imageboards corpus annotated whether a message is aggressive or not. Random forest classifier surpassed CNN for the task of detecting aggression for the English language, however, the gap between two classifiers is very significant for the task of sentiment analysis of movie reviews and Random forest performed poorly for the Russian language. That is why convolutional neural network (and deep learning, in general) classifiers are considered more perspective and promising. We also tried implementing linguistic features, such as part of speech tagging, but it did not lead to better results. However, the results are promising and in future works we will continue their implementation.

\subsubsection*{Acknowledgments.} The survey is being carried out with the support of the Russian Science Foundation (RSF) in the framework of the project № 14-18-01059 at the Institute of Applied and Mathematical Linguistics of the Moscow State Linguistic University and with the support of the Russian Foundation for Basic Research (RFBR) in the framework of the project № 16-29-12986.

\end{document}